%% file: main.tex
\PassOptionsToPackage{dvipsnames}{xcolor}
\documentclass[11pt,letterpaper]{our}
\usepackage{mathalfa}
\usepackage[all]{hypcap}
\usepackage{algorithm}
\usepackage{algorithmicx}
\usepackage{algpseudocode}
\usepackage{graphicx}
\usepackage{subcaption}
\usepackage{booktabs}
\usepackage{float}
\usepackage{bm}
\usepackage{bxcoloremoji}
\usepackage{mathtools}
\usepackage{tikz}
\usepackage{multirow} 
\usepackage{mathrsfs}
\usepackage{nicefrac}
\usepackage{listings}
\usepackage{mathalfa}
\usepackage{dsfont}
\usepackage{fdsymbol}
\usepackage{wrapfig}
\usepackage{lipsum}
\usepackage{stackengine}
\usepackage{adjustbox}
\usepackage{colortbl}  
\usepackage{array}
\usepackage{multicol}
\usepackage{multirow}
\usepackage{gradient-text}
\usepackage{url}
\usepackage{algpseudocode}
\usepackage{fontawesome5}
\usepackage{fancyhdr}
\usepackage{rotating}
\usepackage[capitalise]{cleveref}
\usepackage{listings}

\PassOptionsToPackage{comma,numbers,compress}{natbib}
\usepackage[comma,authoryear,compress]{natbib}
\usepackage[font=small,labelfont=bf]{caption}

\definecolor{lightblue}{rgb}{0.22,0.45,0.70}
\definecolor{YaleBlue}{rgb}{0.06,0.30,0.56}

\newtcolorbox{AIbox}[2][]{aibox,title=#2,#1}

\input{math_commands}

\input{command}

\usepackage{hyperref}
\hypersetup{
    colorlinks = true,
    citecolor = {YaleBlue},
}

\title{\centering \MEMO: Building Production‑Ready AI Agents with Scalable Long‑Term Memory}
\author{\centering Prateek Chhikara, Dev Khant, Saket Aryan, Taranjeet Singh, \textnormal{\footnotesize \textit{and}} Deshraj Yadav\\
\vspace{1em}
\hspace{0.6cm} \href{mailto:research@mem0.ai}{\texttt{research@mem0.ai}}}

\fancypagestyle{firstpage}{
  \fancyhf{}
  \fancyfoot[C]{\thepage}
}

\begin{document}
\input{sections/abs}
\maketitle

\vspace{-31.3em} 

\begin{center}
\includegraphics[width=0.1\textwidth]{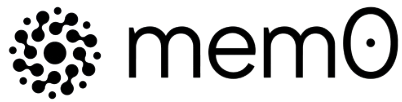}
\end{center}

\vspace{28em}

\input{sections/intro}
\input{sections/proposed_work}

\input{sections/experiment_setup}
\input{sections/result}
\input{sections/conclusion}

\bibliography{cite}
\appendix
\input{sections/appendix}

\end{document}

%% file: math_commands.tex
\usepackage{amsmath,amsfonts,bm}

\def\eqref#1{equation~\ref{#1}}

\def\1{\bm{1}}

\DeclareMathAlphabet{\mathsfit}{\encodingdefault}{\sfdefault}{m}{sl}
\SetMathAlphabet{\mathsfit}{bold}{\encodingdefault}{\sfdefault}{bx}{n}



%% file: command.tex
\definecolor{darkergreen}{RGB}{50,160,50}

\newcommand{\MEMO}{\texttt{Mem0}}

\definecolor{myred}{HTML}{FF8577}
\definecolor{mygreen}{HTML}{0FA958}
\definecolor{myblue}{HTML}{1982C4}
\definecolor{codegreen}{rgb}{0,0.5,0}
\definecolor{codegray}{rgb}{0.5,0.5,0.5}
\definecolor{codepurple}{rgb}{0.07,0,0.53}
\definecolor{codered}{RGB}{189,41,0}
\definecolor{codecomment}{RGB}{153,153,153}
\definecolor{backcolour}{rgb}{0.96,0.96,0.96}
\definecolor{royalblue}{rgb}{0.0, 0.14, 0.4}
\definecolor{egyptianblue}{rgb}{0.06, 0.2, 0.65}
\definecolor{royalazure}{rgb}{0.0, 0.22, 0.66}
\definecolor{portlandorange}{rgb}{1.0, 0.35, 0.21}
\definecolor{sienna}{RGB}{183,105,68}
\definecolor{saddlebrown}{RGB}{139,69,19}
\definecolor{mediumbrown}{RGB}{83,41,11}
\definecolor{darkbrown}{RGB}{58,28,7}

\hypersetup{
    colorlinks=true,
    linkcolor=sienna,
    urlcolor=royalblue,
    citecolor=royalblue,
}

\newcommand{\mem}{\texttt{Mem0}}
\newcommand{\memp}{$\texttt{Mem0}^{\tiny g}$}

\newcommand{\Fone}{F\textsubscript{1}}
\newcommand{\Bone}{B\textsubscript{1}}
\newcommand{\Judge}{J}

%% file: sections/abs.tex
\begin{abstract}

\vspace{1cm}

Large Language Models (LLMs) have demonstrated remarkable prowess in generating contextually coherent responses, yet their fixed context windows pose fundamental challenges for maintaining consistency over prolonged multi-session dialogues. We introduce \mem, a scalable memory-centric architecture that addresses this issue by dynamically extracting, consolidating, and retrieving salient information from ongoing conversations. Building on this foundation, we further propose an enhanced variant that leverages graph-based memory representations to capture complex relational structures among conversational elements.
Through comprehensive evaluations on the \texttt{LOCOMO} benchmark, we systematically compare our approaches against six baseline categories: (i) established memory-augmented systems, (ii) retrieval-augmented generation (RAG) with varying chunk sizes and
$k$-values, (iii) a full-context approach that processes the entire conversation history, (iv) an open-source memory solution, (v) a proprietary model system, and (vi) a dedicated memory management platform. 
Empirical results demonstrate that our methods consistently outperform all existing memory systems across four question categories: single-hop, temporal, multi-hop, and open-domain. 
Notably, \mem~achieves 26\% relative improvements in the LLM-as-a-Judge metric over OpenAI, while \mem~with graph memory achieves around 2\% higher overall score than the base \mem~configuration.
Beyond accuracy gains, we also markedly reduce computational overhead compared to the full-context approach. In particular, \mem~attains a 91\% lower p95 latency and saves more than 90\% token cost, thereby offering a compelling balance between advanced reasoning capabilities and practical deployment constraints.
Our findings highlight the critical role of structured, persistent memory mechanisms for long-term conversational coherence, paving the way for more reliable and efficient LLM-driven AI agents.

\vspace{5mm}
\textbf{Code can be found at}: \href{https://mem0.ai/research}{https://mem0.ai/research}

\end{abstract}

%% file: sections/intro.tex
\section{Introduction}
\label{sec:intro}

\begin{figure}[!t]
    \centering
    \begin{minipage}[t]{0.65\textwidth}
        \centering
        \includegraphics[width=\textwidth]{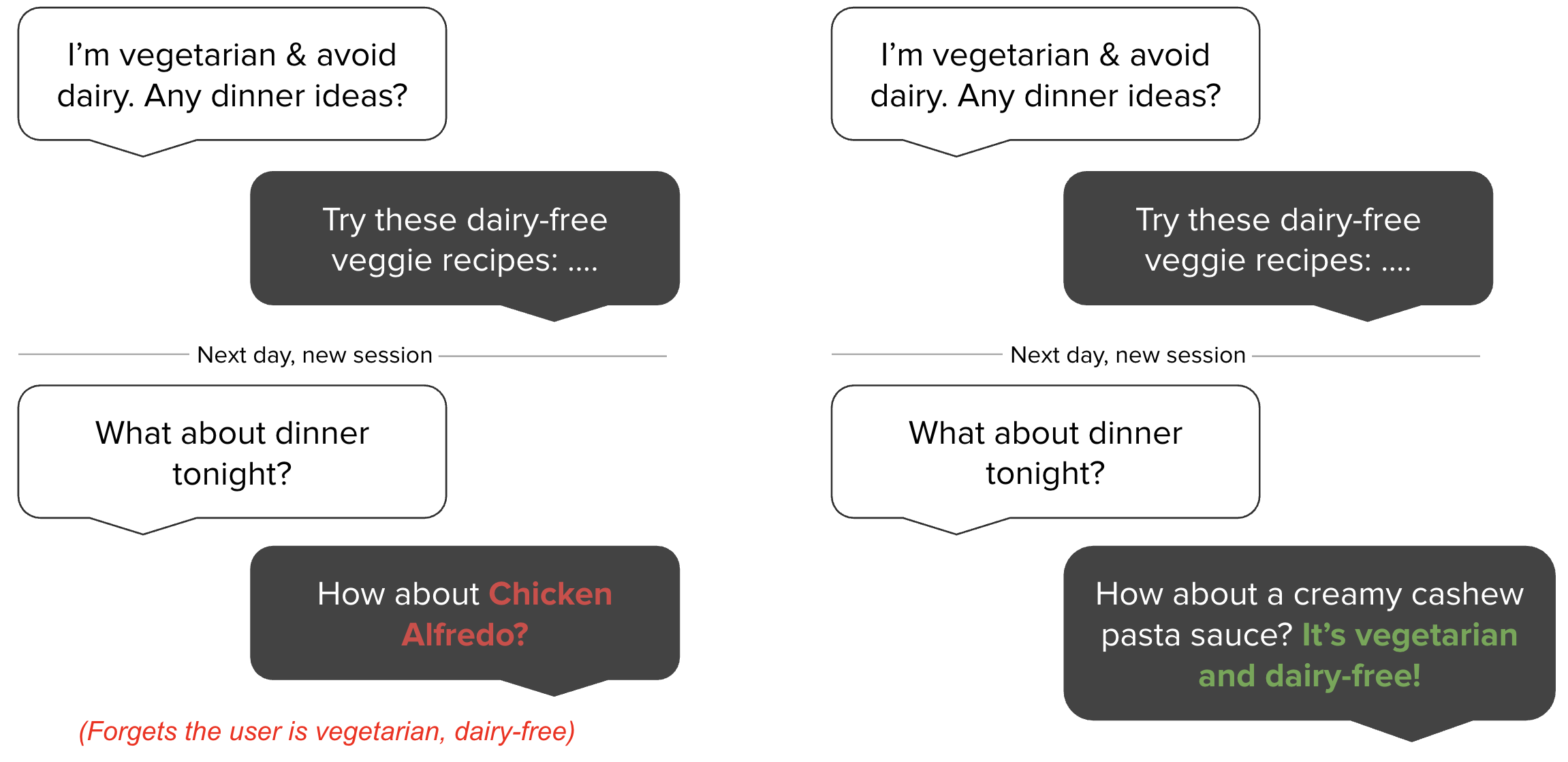}
    \end{minipage}%
    \hfill
    \begin{minipage}[t]{0.32\textwidth}
        \vspace{-5.3cm}
        \captionof{figure}{\textbf{Illustration of memory importance in AI agents.} 
        \textit{Left}: Without persistent memory, the system forgets critical user information (vegetarian, dairy-free preferences) between sessions, resulting in inappropriate recommendations. 
        \textit{Right}: With effective memory, the system maintains these dietary preferences across interactions, enabling contextually appropriate suggestions that align with previously established constraints.}
        \label{fig:main}
    \end{minipage}
\end{figure}

Human memory is a \emph{foundation of intelligence}—it shapes our identity, guides decision-making, and enables us to learn, adapt, and form meaningful relationships \citep{craik1992human}. Among its many roles, memory is essential for communication: we recall past interactions, infer preferences, and construct evolving mental models of those we engage with \citep{assmann2011communicative}. This ability to retain and retrieve information over extended periods enables coherent, contextually rich exchanges that span days, weeks, or even months. AI agents, powered by large language models (LLMs), have made remarkable progress in generating fluent, contextually appropriate responses \citep{yu2024finmem, zhang2024survey}. However, these systems are fundamentally limited by their reliance on fixed context windows, which severely restrict their ability to maintain coherence over extended interactions \citep{bulatov2022recurrent, liu2023think}. 
This limitation stems from LLMs' lack of persistent memory mechanisms that can extend beyond their finite context windows. While humans naturally accumulate and organize experiences over time, forming a continuous narrative of interactions, AI systems cannot inherently persist information across separate sessions or after context overflow.
The absence of persistent memory creates a fundamental disconnect in human-AI interaction. Without memory, AI agents forget user preferences, repeat questions, and contradict previously established facts. 
Consider a simple example illustrated in Figure \ref{fig:main}, where a user mentions being vegetarian and avoiding dairy products in an initial conversation. 
In a subsequent session, when the user asks about dinner recommendations, a system without persistent memory might suggest chicken, completely contradicting the established dietary preferences. In contrast, a system with persistent memory would maintain this critical user information across sessions and suggest appropriate vegetarian, dairy-free options. This common scenario highlights how memory failures can fundamentally undermine user experience and trust.

Beyond conversational settings, memory mechanisms have been shown to dramatically enhance agent performance in interactive environments \citep{majumderclin, shinn2023reflexion}. Agents equipped with memory of past experiences can better anticipate user needs, learn from previous mistakes, and generalize knowledge across tasks \citep{chhikara2023knowledge}. Research demonstrates that memory-augmented agents improve decision-making by leveraging causal relationships between actions and outcomes, leading to more effective adaptation in dynamic scenarios \citep{rasmussen2025zep}. Hierarchical memory architectures \citep{packer2023memgpt, sarthi2024raptor} and agentic memory systems capable of autonomous evolution \citep{xu2025mem} have further shown that memory enables more coherent, long-term reasoning across multiple dialogue sessions.

Unlike humans, who dynamically integrate new information and revise outdated beliefs, LLMs effectively ``\textit{reset}" once information falls outside their context window \citep{zhang2024guided, timoneda2025memory}. 
Even as models like OpenAI's GPT-4 (128K tokens) \citep{hurst2024gpt}, o1 (200K context) \citep{jaech2024openai}, Anthropic's Claude 3.7 Sonnet (200K tokens) \citep{anthropic2023modelcard}, and Google's Gemini (at least 10M tokens) \citep{team2024gemini} push the boundaries of context length, these improvements merely delay rather than solve the fundamental limitation.
 In practical applications, even these extended context windows prove insufficient for two critical reasons. First, as meaningful human-AI relationships develop over weeks or months, conversation history inevitably exceeds even the most generous context limits. Second, and perhaps more importantly, real-world conversations rarely maintain thematic continuity. A user might mention dietary preferences (being vegetarian), then engage in hours of unrelated discussion about programming tasks, before returning to food-related queries about dinner options. In such scenarios, a full-context approach would need to reason through mountains of irrelevant information, with the critical dietary preferences potentially buried among thousands of tokens of coding discussions. Moreover, simply presenting longer contexts does not ensure effective retrieval or utilization of past information, as attention mechanisms degrade over distant tokens \citep{guo2024active, nelson2024needle}.
 This limitation is particularly problematic in high-stakes domains such as healthcare, education, and enterprise support, where maintaining continuity and trust is crucial \citep{hatalis2023memory}. To address these challenges, AI agents must adopt memory systems that go beyond static context extension. A robust AI memory should selectively store important information, consolidate related concepts, and retrieve relevant details when needed—\emph{mirroring human cognitive processes} \citep{he2024human}. By integrating such mechanisms, we can develop AI agents that maintain consistent personas, track evolving user preferences, and build upon prior exchanges. This shift will transform AI from transient, forgetful responders into reliable, long-term collaborators, fundamentally redefining the future of conversational intelligence.

In this paper, we address a fundamental limitation in AI systems: their inability to maintain coherent reasoning across extended conversations across different sessions, which severely restricts meaningful long-term interactions with users. We introduce \mem~(pronounced as \emph{mem-zero}), a novel memory architecture that dynamically captures, organizes, and retrieves salient information from ongoing conversations. Building on this foundation, we develop \memp, which enhances the base architecture with graph-based memory representations to better model complex relationships between conversational elements. 
Our experimental results on the \texttt{LOCOMO} benchmark demonstrate that our approaches consistently outperform existing memory systems—including memory-augmented architectures, retrieval-augmented generation (RAG) methods, and both open-source and proprietary solutions—across diverse question types, while simultaneously requiring significantly lower computational resources.
Latency measurements further reveal that \mem~operates with 91\% lower response times than full-context approaches, striking an optimal balance between sophisticated reasoning capabilities and practical deployment constraints. These contributions represent a meaningful step toward AI systems that can maintain coherent, context-aware conversations over extended durations—mirroring human communication patterns and opening new possibilities for applications in personal tutoring, healthcare, and personalized assistance.

%% file: sections/proposed_work.tex
\section{Proposed Methods}
\label{sec:proposed_work}

We introduce two memory architectures for AI agents. \textbf{(1)} \mem~implements a novel paradigm that extracts, evaluates, and manages salient information from conversations through dedicated modules for memory extraction and updation. The system processes a pair of messages between either two user participants or a user and an assistant. \textbf{(2)} \memp~extends this foundation by incorporating graph-based memory representations, where memories are stored as directed labeled graphs with entities as nodes and relationships as edges. This structure enables a deeper understanding of the connections between entities. By explicitly modeling both entities and their relationships, \memp~supports more advanced reasoning across interconnected facts, especially for queries that require navigating complex relational paths across multiple memories.

\subsection{\mem}
\begin{figure}[!t]
    \centering
    \includegraphics[width=\textwidth]{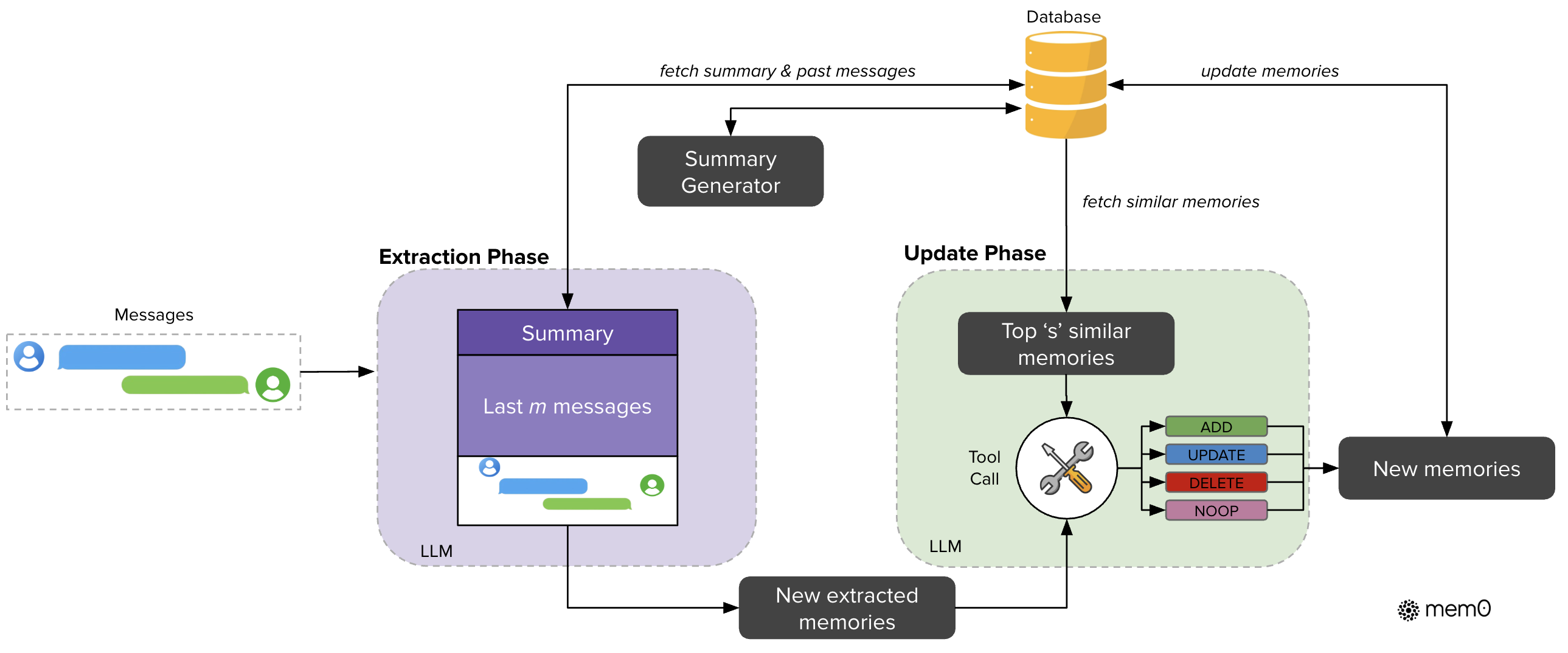}
    \caption{Architectural overview of the \mem~system showing extraction and update phase. The extraction phase processes messages and historical context to create new memories. The update phase evaluates these extracted memories against similar existing ones, applying appropriate operations through a Tool Call mechanism. The database serves as the central repository, providing context for processing and storing updated memories.}
    \label{fig:mem0}
\end{figure}

Our architecture follows an incremental processing paradigm, enabling it to operate seamlessly within ongoing conversations. As illustrated in Figure \ref{fig:mem0}, the complete pipeline architecture consists of two phases: \textit{extraction} and \textit{update}.

The \textbf{extraction phase} initiates upon ingestion of a new message pair $(m_{t-1}, m_t)$, where $m_t$ represents the current message and $m_{t-1}$ the preceding one. This pair typically consists of a user message and an assistant response, capturing a complete interaction unit. To establish appropriate context for memory extraction, the system employs two complementary sources: \texttt{(1)} a conversation summary $S$ retrieved from the database that encapsulates the semantic content of the entire conversation history, and \texttt{(2)} a sequence of recent messages $\{m_{t-m}, m_{t-m+1}, ..., m_{t-2}\}$ from the conversation history, where $m$ is a hyperparameter controlling the recency window. To support context-aware memory extraction, we implement an asynchronous summary generation module that periodically refreshes the conversation summary. This component operates independently of the main processing pipeline, ensuring that memory extraction consistently benefits from up-to-date contextual information without introducing processing delays. While $S$ provides global thematic understanding across the entire conversation, the recent message sequence offers granular temporal context that may contain relevant details not consolidated in the summary. This dual contextual information, combined with the new message pair, forms a comprehensive prompt $P = (S, \{m_{t-m}, ..., m_{t-2}\}, m_{t-1}, m_t)$ for an extraction function $\phi$ implemented via an LLM. The function $\phi(P)$ then extracts a set of salient memories $\Omega = \{\omega_1, \omega_2, ..., \omega_n\}$ specifically from the new exchange while maintaining awareness of the conversation's broader context, resulting in candidate facts for potential inclusion in the knowledge base. 

Following extraction, the \textbf{update phase} evaluates each candidate fact against existing memories to maintain consistency and avoid redundancy. This phase determines the appropriate memory management operation for each extracted fact $\omega_i \in \Omega$. Algorithm \ref{alg:memory_update}, mentioned in Appendix \ref{appendix:algorithm}, illustrates this process. For each fact, the system first retrieves the top $s$ semantically similar memories using vector embeddings from the database. These retrieved memories, along with the candidate fact, are then presented to the LLM through a function-calling interface we refer to as a `tool call.' The LLM itself determines which of four distinct operations to execute: \texttt{ADD} for creation of new memories when no semantically equivalent memory exists; \texttt{UPDATE} for augmentation of existing memories with complementary information; \texttt{DELETE} for removal of memories contradicted by new information; and \texttt{NOOP} when the candidate fact requires no modification to the knowledge base. Rather than using a separate classifier, we leverage the LLM's reasoning capabilities to directly select the appropriate operation based on the semantic relationship between the candidate fact and existing memories. Following this determination, the system executes the provided operations, thereby maintaining knowledge base coherence and temporal consistency.

In our experimental evaluation, we configured the system with `$m$' = 10 previous messages for contextual reference and `$s$' = 10 similar memories for comparative analysis. All language model operations utilized \texttt{GPT-4o-mini} as the inference engine. The vector database employs dense embeddings to facilitate efficient similarity search during the update phase.

\subsection{\memp}
\begin{figure}[!t]
    \centering
    \includegraphics[width=\textwidth]{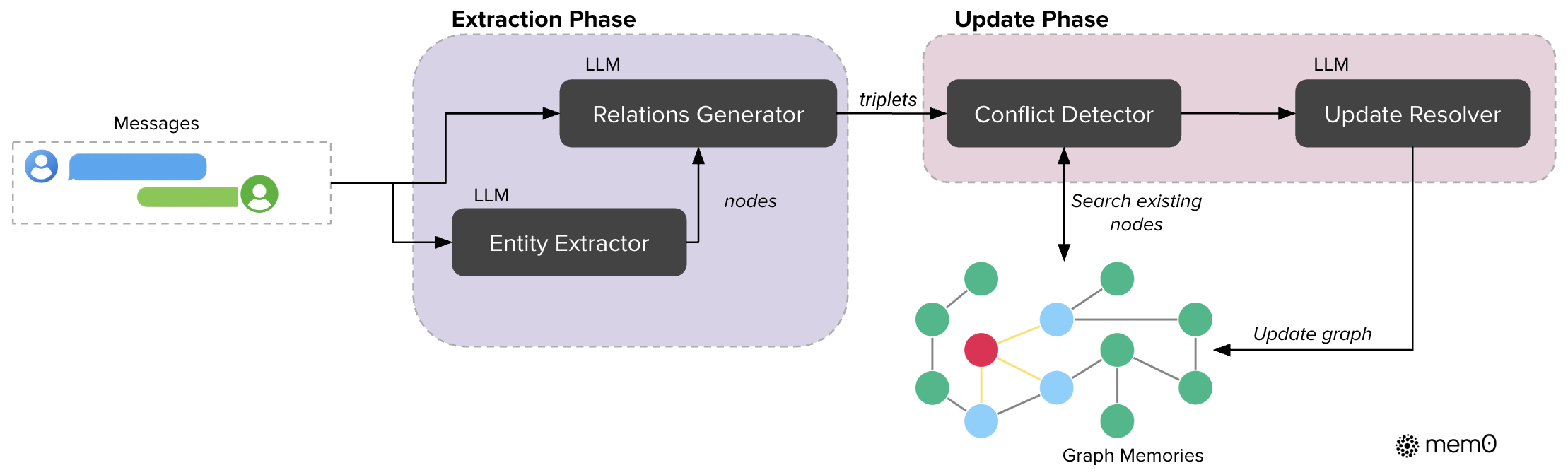}
    \caption{Graph-based memory architecture of \memp~illustrating entity extraction and update phase. The extraction phase uses LLMs to convert conversation messages into entities and relation triplets. The update phase employs conflict detection and resolution mechanisms when integrating new information into the existing knowledge graph.}
    \label{fig:mem0p}
\end{figure}

The \memp~pipeline, illustrated in Figure \ref{fig:mem0p}, implements a graph-based memory approach that effectively captures, stores, and retrieves contextual information from natural language interactions \citep{zhang2022study}. In this framework, memories are represented as a directed labeled graph $G = (V, E, L)$, where:

\begin{itemize}
    \item Nodes $V$ represent entities (e.g., \textsc{\textcolor[HTML]{FF9900}{Alice}}, \textsc{\textcolor[HTML]{FF9900}{San\_Francisco}})
    \item Edges $E$ represent relationships between entities (e.g., \textsc{\textcolor[HTML]{0099CC}{lives\_in}})
    \item Labels $L$ assign semantic types to nodes (e.g., \textsc{Alice} - \textcolor[HTML]{009933}{Person}, \textsc{San\_Francisco} - \textcolor[HTML]{009933}{City})
\end{itemize}

Each entity node $v \in V$ contains three components: (1) an entity type classification that categorizes the entity (e.g., Person, Location, Event), (2) an embedding vector $e_v$ that captures the entity's semantic meaning, and (3) metadata including a creation timestamp $t_v$. Relationships in our system are structured as triplets in the form $(v_s, r, v_d)$, where $v_s$ and $v_d$ are source and destination entity nodes, respectively, and $r$ is the labeled edge connecting them.

The extraction process employs a two-stage pipeline leveraging LLMs to transform unstructured text into structured graph representations. First, an \textit{\textbf{entity extractor}} module processes the input text to identify a set of entities along with their corresponding types. In our framework, entities represent the key information elements in conversations—including people, locations, objects, concepts, events, and attributes that merit representation in the memory graph. The entity extractor identifies these diverse information units by analyzing the semantic importance, uniqueness, and persistence of elements in the conversation. For instance, in a conversation about travel plans, entities might include destinations (cities, countries), transportation modes, dates, activities, and participant preferences—essentially any discrete information that could be relevant for future reference or reasoning. 

Next, a \textit{\textbf{relationship generator}} component derives meaningful connections between these entities, establishing a set of relationship triplets that capture the semantic structure of the information. This LLM-based module analyzes the extracted entities and their context within the conversation to identify semantically significant connections. It works by examining linguistic patterns, contextual cues, and domain knowledge to determine how entities relate to one another. For each potential entity pair, the generator evaluates whether a meaningful relationship exists and, if so, classifies this relationship with an appropriate label (e.g., `lives\_in', `prefers', `owns', `happened\_on'). The module employs prompt engineering techniques that guide the LLM to reason about both explicit statements and implicit information in the dialogue, resulting in relationship triplets that form the edges in our memory graph and enable complex reasoning across interconnected information.
When integrating new information, \memp~employs a sophisticated storage and update strategy. For each new relationship triple, we compute embeddings for both source and destination entities, then search for existing nodes with semantic similarity above a defined threshold `$t$'. Based on node existence, the system may create both nodes, create only one node, or use existing nodes before establishing the relationship with appropriate metadata. To maintain a consistent knowledge graph, we implement a \textit{\textbf{conflict detection}} mechanism that identifies potentially conflicting existing relationships when new information arrives. An LLM-based \textbf{\textit{update resolver}} determines if certain relationships should be obsolete, marking them as invalid rather than physically removing them to enable temporal reasoning.

The memory retrieval functionality in \memp~implements a dual-approach strategy for optimal information access. The entity-centric method first identifies key entities within a query, then leverages semantic similarity to locate corresponding nodes in the knowledge graph. It systematically explores both incoming and outgoing relationships from these anchor nodes, constructing a comprehensive subgraph that captures relevant contextual information. Complementing this, the semantic triplet approach takes a more holistic view by encoding the entire query as a dense embedding vector. This query representation is then matched against textual encodings of each relationship triplet in the knowledge graph. The system calculates fine-grained similarity scores between the query and all available triplets, returning only those that exceed a configurable relevance threshold, ranked in order of decreasing similarity. This dual retrieval mechanism enables \memp~to handle both targeted entity-focused questions and broader conceptual queries with equal effectiveness.

From an implementation perspective, the system utilizes Neo4j\footnote{\url{https://neo4j.com/}} as the underlying graph database. LLM-based extractors and update module leverage \texttt{GPT-4o-mini} with function calling capabilities, allowing for structured extraction of information from unstructured text. By combining graph-based representations with semantic embeddings and LLM-based information extraction, \memp~achieves both the structural richness needed for complex reasoning and the semantic flexibility required for natural language understanding.

%% file: sections/experiment_setup.tex
\section{Experimental Setup}
\label{sec:experiment_setup}

\subsection{Dataset}
The \texttt{LOCOMO} \citep{maharana2024evaluating} dataset is designed to evaluate long-term conversational memory in dialogue systems. It comprises 10 extended conversations, each containing approximately 600 dialogues and 26000 tokens on average, distributed across multiple sessions. Each conversation captures two individuals discussing daily experiences or past events. Following these multi-session dialogues, each conversation is accompanied by 200 questions on an average with corresponding ground truth answers. These questions are categorized into multiple types: single-hop, multi-hop, temporal, and open-domain. The dataset originally included an adversarial question category, which was designed to test systems' ability to recognize unanswerable questions. However, this category was excluded from our evaluation because ground truth answers were unavailable, and the expected behavior for this question type is that the agent should recognize them as unanswerable.

\subsection{Evaluation Metrics}
Our evaluation framework implements a comprehensive approach to assess long-term memory capabilities in dialogue systems, considering both response quality and operational efficiency. We categorize our metrics into two distinct groups that together provide a holistic understanding of system performance.

\paragraph{\texttt{(1)} Performance Metrics}
Previous research in conversational AI \citep{goswami2025dissecting, soni2024evaluating, singh2020ensemble} has predominantly relied on lexical similarity metrics such as \textbf{F1 Score} (\Fone) and \textbf{BLEU-1} (\Bone). However, these metrics exhibit significant limitations when evaluating factual accuracy in conversational contexts. Consider a scenario where the ground truth answer is `\textit{Alice was born in March}' and a system generates `\textit{Alice is born in July}.' Despite containing a critical factual error regarding the birth month, traditional metrics would assign relatively high scores due to lexical overlap in the remaining tokens (`\textit{Alice},' `\textit{born},' etc.). This fundamental limitation can lead to misleading evaluations that fail to capture semantic correctness.
To address these shortcomings, we use \textbf{LLM-as-a-Judge} (\Judge) as a complementary evaluation metric. This approach leverages a separate, more capable LLM to assess response quality across multiple dimensions, including factual accuracy, relevance, completeness, and contextual appropriateness. The judge model analyzes the question, ground truth answer and the generated answer, providing a more nuanced evaluation that aligns better with human judgment. Due to the stochastic nature of \Judge~evaluations, we conducted 10 independent runs for each method on the entire dataset and report the mean scores along with $\pm$1 standard deviation. More details about the \Judge~is present in Appendix \ref{appendix:llm_judge}.

\paragraph{\texttt{(2)} Deployment Metrics}
Beyond response quality, practical deployment considerations are crucial for real-world applications of long-term memory in AI agents. We systematically track \textbf{Token Consumption}, using `\texttt{cl100k\_base}' encoding from \texttt{tiktoken}, measuring the number of tokens extracted during retrieval that serve as context for answering queries. For our memory-based models, these tokens represent the memories retrieved from the knowledge base, while for RAG-based models, they correspond to the total number of tokens in the retrieved text chunks. This distinction is important as it directly affects operational costs and system efficiency—whether processing concise memory facts or larger raw text segments. We further monitor \textbf{Latency}, (i) \emph{search latency}: which captures the total time required to search the memory (in memory-based solutions) or chunk (in RAG-based solutions) and (ii) \emph{total latency:} time to generate appropriate responses, consisting of both retrieval time (accessing memories or chunks) and answer generation time using the LLM.

The relationship between these metrics reveals important trade-offs in system design. For instance, more sophisticated memory architectures might achieve higher factual accuracy but at the cost of increased token consumption and latency. Our multi-dimensional evaluation methodology enables researchers and practitioners to make informed decisions based on their specific requirements, whether prioritizing response quality for critical applications or computational efficiency for real-time deployment scenarios. 

\subsection{Baselines}
To comprehensively evaluate our approach, we compare against six distinct categories of baselines that represent the current state of conversational memory systems. These diverse baselines collectively provide a robust framework for evaluating the effectiveness of different memory architectures across various dimensions, including factual accuracy, computational efficiency, and scalability to extended conversations. Where applicable, unless otherwise specified, we set the temperature to 0 to ensure the runs are as reproducible as possible.

\paragraph{Established \texttt{LOCOMO} Benchmarks}
We first establish a comparative foundation by evaluating previously benchmarked methods on the \texttt{LOCOMO} dataset. These include five established approaches: LoCoMo \citep{maharana2024evaluating}, ReadAgent \citep{lee2024human}, MemoryBank \citep{zhong2024memorybank}, MemGPT \citep{packer2023memgpt}, and A-Mem \citep{xu2025mem}. These established benchmarks not only provide direct comparison points with published results but also represent the evolution of conversational memory architectures across different algorithmic paradigms. For our evaluation, we select the metrics where \texttt{gpt-4o-mini} was used for the evaluation. More details about these benchmarks are mentioned in Appendix \ref{appendix:baselines}.

\paragraph{Open-Source Memory Solutions}
Our second category consists of promising open-source memory architectures such as LangMem\footnote{\url{https://langchain-ai.github.io/langmem/}} (Hot Path) that have demonstrated effectiveness in related conversational tasks but have not yet been evaluated on the \texttt{LOCOMO} dataset. By adapting these systems to our evaluation framework, we broaden the comparative landscape and identify potential alternative approaches that may offer competitive performance. We initialized the LLM with \texttt{gpt-4o-mini} and used \texttt{text-embedding-small-3} as the embedding model.

\paragraph{Retrieval-Augmented Generation (RAG)}
As a baseline, we treat the entire conversation history as a document collection and apply a standard RAG pipeline. We first segment each conversation into fixed‐length chunks (128, 256, 512, 1024, 2048, 4096, and 8192 tokens), where 8192 is the maximum chunk size supported by our embedding model. All chunks are embedded using OpenAI’s \texttt{text-embedding-small-3} to ensure consistent vector quality across configurations. At query time, we retrieve the top $k$ chunks by semantic similarity and concatenate them as context for answer generation. Throughout our experiments we set $k$$\in$\{1,2\}: with $k$=1 only the single most relevant chunk is used, and with $k$=2 the two most relevant chunks (up to 16384 tokens) are concatenated. We avoid \(k>2\) since the average conversation length (26000 tokens) would be fully covered, negating the benefits of selective retrieval. By varying chunk size and \(k\), we systematically evaluate RAG performance on long‐term conversational memory tasks.

\paragraph{Full-Context Processing}
We adopt a straightforward approach by passing the entire conversation history within the context window of the LLM. This method leverages the model's inherent ability to process sequential information without additional architectural components. While conceptually simple, this approach faces practical limitations as conversation length increases, eventually increasing token cost and latency. Nevertheless, it establishes an important reference point for understanding the value of more sophisticated memory mechanisms compared to direct processing of available context.

\paragraph{Proprietary Models} We evaluate OpenAI's memory\footnote{\url{https://openai.com/index/memory-and-new-controls-for-chatgpt/}} feature available in their ChatGPT interface, specifically using \texttt{gpt-4o-mini} for consistency. We ingest entire \texttt{LOCOMO} conversations with a prompt (see Appendix \ref{appendix:llm_judge}) into single chat sessions, prompting memory generation with timestamps, participant names, and conversation text. These generated memories are then used as complete context for answering questions about each conversation, intentionally granting the OpenAI approach privileged access to all memories rather than only question-relevant ones. This methodology accommodates the lack of external API access for selective memory retrieval in OpenAI's system for benchmarking.

\paragraph{Memory Providers} We incorporate Zep \citep{rasmussen2025zep}, a memory management platform designed for AI agents. Using their platform version, we conduct systematic evaluations across the \texttt{LOCOMO} dataset, maintaining temporal fidelity by preserving timestamp information alongside conversational content. This temporal anchoring ensures that time-sensitive queries can be addressed through appropriately contextualized memory retrieval, particularly important for evaluating questions that require chronological awareness. This baseline represents an important commercial implementation of memory management specifically engineered for AI agents.

\begin{table*}[!t]
\centering
\caption{Performance comparison of memory-enabled systems across different question types in the \texttt{LOCOMO} dataset. Evaluation metrics include F1 score (\Fone), BLEU-1 (\Bone), and LLM-as-a-Judge score (\Judge), with higher values indicating better performance. $\text{A-Mem}^{*}$ represents results from our re-run of A-Mem to generate LLM-as-a-Judge scores by setting temperature as 0. \memp~indicates our proposed architecture enhanced with graph memory. \textbf{Bold} denotes the best performance for each metric across all methods. ($\uparrow$) represents higher score is better.}
\label{tab:results}
\resizebox{\textwidth}{!}{%
\begin{tabular}{l|ccc|ccc|ccc|ccc}
\toprule
\multirow{2}{*}{\textbf{Method}}
  & \multicolumn{3}{c|}{\textbf{Single Hop}}
  & \multicolumn{3}{c|}{\textbf{Multi-Hop}}
  & \multicolumn{3}{c|}{\textbf{Open Domain}}
  & \multicolumn{3}{c}{\textbf{Temporal}} \\
  & \Fone~$\uparrow$ & \Bone~$\uparrow$ & \Judge~$\uparrow$
  & \Fone~$\uparrow$ & \Bone~$\uparrow$ & \Judge~$\uparrow$
  & \Fone~$\uparrow$ & \Bone~$\uparrow$ & \Judge~$\uparrow$
  & \Fone~$\uparrow$ & \Bone~$\uparrow$ & \Judge~$\uparrow$ \\
\midrule
LoCoMo      & 25.02 & 19.75 & --    & 12.04 & 11.16 & --    & 40.36 & 29.05 & --    & 18.41 & 14.77 & --    \\
ReadAgent   &  9.15 &  6.48 & --    &  5.31 &  5.12 & --    &  9.67 &  7.66 & --    & 12.60 &  8.87 & --    \\
MemoryBank  &  5.00 &  4.77 & --    &  5.56 &  5.94 & --    &  6.61 &  5.16 & --    &  9.68 &  6.99 & --    \\
MemGPT      & 26.65 & 17.72 & --    &  9.15 &  7.44 & --    & 41.04 & 34.34 & --    & 25.52 & 19.44 & --    \\
A-Mem       & 27.02 & 20.09 & --    & 12.14 & 12.00 & --    & 44.65 & 37.06 & --    & 45.85 & 36.67 & --    \\
A-Mem*      & 20.76 & 14.90 & 39.79 ± 0.38 &  9.22 &  8.81 & 18.85 ± 0.31 & 33.34 & 27.58 & 54.05 ± 0.22 & 35.40 & 31.08 & 49.91 ± 0.31 \\
LangMem     & 35.51 & 26.86 & 62.23 ± 0.75 & 26.04 & \textbf{22.32} & 47.92 ± 0.47 & 40.91 & 33.63 & 71.12 ± 0.20 & 30.75 & 25.84 & 23.43 ± 0.39 \\
Zep         & 35.74 & 23.30 & 61.70 ± 0.32 & 19.37 & 14.82 & 41.35 ± 0.48 & \textbf{49.56} & 38.92 & \textbf{76.60 ± 0.13} & 42.00 & 34.53 & 49.31 ± 0.50 \\
OpenAI      & 34.30 & 23.72 & 63.79 ± 0.46 & 20.09 & 15.42 & 42.92 ± 0.63 & 39.31 & 31.16 & 62.29 ± 0.12 & 14.04 & 11.25 & 21.71 ± 0.20 \\
\midrule
\mem        & \textbf{38.72} & \textbf{27.13} & \textbf{67.13 ± 0.65}
            & \textbf{28.64} & 21.58 & \textbf{51.15 ± 0.31}
            & 47.65 & 38.72 & 72.93 ± 0.11
            & 48.93 & \textbf{40.51} & 55.51 ± 0.34 \\
\memp       & 38.09 & 26.03 & 65.71 ± 0.45
            & 24.32 & 18.82 & 47.19 ± 0.67
            & 49.27 & \textbf{40.30} & 75.71 ± 0.21
            & \textbf{51.55} & 40.28 & \textbf{58.13 ± 0.44} \\
\bottomrule
\end{tabular}%
}
\end{table*}

%% file: sections/result.tex
\section{Evaluation Results, Analysis and Discussion.}

\subsection{Performance Comparison Across Memory-Enabled Systems}
Table~\ref{tab:results} reports \Fone, \Bone~and \Judge~scores for our two architectures—\mem~and \memp~—against a suite of competitive baselines, as mentioned in Section \ref{sec:experiment_setup}, on single‑hop, multi‑hop, open‑domain, and temporal questions. Overall, both of our models set new state‑of‑the‑art marks in all the three evaluation metrics for most question types.

\paragraph{Single-Hop Question Performance} Single‑hop queries involve locating a single factual span contained within one dialogue turn. Leveraging its dense memories in natural language text, \mem~secures the strongest results:\Fone=38.72, \Bone=27.13, and \Judge=67.13. Augmenting the natural language memories with graph memory (\memp) yields marginal performance drop compared to \mem, indicating that relational structure provides limited utility when the retrieval target occupies a single turn. Among the existing baselines, the full‑context \texttt{OpenAI} run attains the next‑best \Judge\ score, reflecting the benefits of retaining the entire conversation in context, while LangMem and Zep both score around 8\% relatively less against our models on \Judge~score. Previous \texttt{LOCOMO} benchmarks such as A-mem lag by more than 25 points in \Judge, underscoring the necessity of fine‑grained, structured memory indexing even for simple retrieval tasks.

\paragraph{Multi-Hop Question Performance}
Multi-hop queries require synthesizing information dispersed across multiple conversation sessions, posing significant challenges in memory integration and retrieval. \mem~clearly outperforms other methods with an \Fone~score of 28.64 and a \Judge~score of 51.15, reflecting its capability to efficiently retrieve and integrate disparate information stored across sessions. Interestingly, the addition of graph memory in \memp~does not provide performance gains here, indicating potential inefficiencies or redundancies in structured graph representations for complex integrative tasks compared to dense natural language memory alone. Baselines like LangMem show competitive performances, but their scores substantially trail those of \mem, emphasizing the advantage of our refined memory indexing and retrieval mechanisms for complex query processing.

\paragraph{Open-Domain Performance}
In open-domain settings, the baseline Zep achieves the highest \Fone\ (49.56) and \Judge\ (76.60) scores, edging out our methods by a narrow margin.  In particular, Zep’s \Judge\ score of 76.60 surpasses \memp’s 75.71 by just 0.89 percentage points and outperforms \mem’s 72.93 by 3.67 points, highlighting a consistent, if slight, advantage in integrating conversational memory with external knowledge. \memp remains a strong runner-up, with a \Judge\ of 75.71 reflecting high factual retrieval precision, while \mem~follows with 72.93, demonstrating robust coherence. These results underscore that although structured relational memories (as in \mem\ and \memp) substantially improve open-domain retrieval, Zep maintains a small but meaningful lead.

\paragraph{Temporal Reasoning Performance}
Temporal reasoning tasks hinge on accurate modeling of event sequences, their relative ordering, and durations within conversational history. Our architectures demonstrate substantial improvements across all metrics, with \memp~achieving the highest \Fone (51.55) and \Judge~(58.13), suggesting that structured relational representations in addition to natural language memories significantly aid in temporally grounded judgments. Notably, the base variant, \mem, also provide a decent \Judge~score (55.51), suggesting that natural language alone can aid in temporally grounded judgments. 
Among baselines, OpenAI notably underperforms, with scores below 15\%, primarily due to missing timestamps in most generated memories despite explicit prompting in the OpenAI ChatGPT to extract memories with timestamps. Other baselines such as A-Mem achieve respectable results, yet our models clearly advance the state-of-the-art, emphasizing the critical advantage of accurately leveraging both natural language contextualization and structured graph representations for temporal reasoning.

\subsection{Cross-Category Analysis}
The comprehensive evaluation across diverse question categories reveals that our proposed architectures, \mem~and \memp, consistently achieve superior performance compared to baseline systems. For single-hop queries, \mem~demonstrates particularly strong performance, benefiting from its efficient dense natural language memory structure. Although graph-based representations in \memp~slightly lag behind in lexical overlap metrics for these simpler queries, they significantly enhance semantic coherence, as demonstrated by competitive \Judge~scores. This indicates that graph structures are more beneficial in scenarios involving nuanced relational context rather than straightforward retrieval. For multi-hop questions, \mem~exhibits clear advantages by effectively synthesizing dispersed information across multiple sessions, confirming that natural language memories provide sufficient representational richness for these integrative tasks. Surprisingly, the expected relational advantages of \memp~do not translate into better outcomes here, suggesting potential overhead or redundancy when navigating more intricate graph structures in multi-step reasoning scenarios.

\begin{table}[!t]
\centering
\caption{Performance comparison of various baselines with proposed methods. Latency measurements show p50 (median) and p95 (95th percentile) values in seconds for both search time (time taken to fetch memories/chunks) and total time (time to generate the complete response). Overall LLM-as-a-Judge score ($\mathrm{J}$) represents the quality metric of the generated responses on the entire \texttt{LOCOMO} dataset.}
\begin{tabular*}{\textwidth}{@{\extracolsep{\fill}}lcccccccc@{}}
\toprule
\multirow{3}{*}{\textit{\textbf{Method}}} & \multicolumn{2}{c}{} & \multicolumn{4}{c}{\textit{\textbf{Latency (seconds)}}} & \multirow{3}{*}{\begin{tabular}[c]{@{}c@{}}\textit{\textbf{Overall}} \\$\mathrm{J}$\end{tabular}} \\
\cmidrule(lr){4-7}
 & \multicolumn{2}{c}{} & \multicolumn{2}{c}{\textbf{Search}} & \multicolumn{2}{c}{\textbf{Total}} & \\
\cmidrule(lr){4-5} \cmidrule(lr){6-7}
& \textbf{K}
& \begin{tabular}{@{}c@{}}
    \textbf{chunk size} / \\
    \textbf{memory tokens}
  \end{tabular}
& \textbf{p50} & \textbf{p95}
& \textbf{p50} & \textbf{p95}
& \\
\midrule
\multirow{14}{*}{RAG} & \multirow{7}{*}{1} & 128 & 0.281  & 0.823 & 0.774 & 1.825 & 47.77 $\pm$ 0.23\% \\
 &  & 256 & 0.251 & 0.710 & 0.745 & 1.628 & 50.15 $\pm$ 0.16\% \\
 &  & 512 & 0.240 & 0.639 & 0.772 & 1.710 & 46.05 $\pm$ 0.14\% \\
 &  & 1024 & 0.240 & 0.723 & 0.821 & 1.957 & 40.74  $\pm$ 0.17\% \\
 &  & 2048 & 0.255 & 0.752 & 0.996 & 2.182 & 37.93 $\pm$ 0.12\% \\
 &  & 4096 & 0.254 & 0.719 & 1.093 & 2.711 & 36.84 $\pm$ 0.17\% \\
 &  & 8192 & 0.279 & 0.838 & 1.396 & 4.416 & 44.53 $\pm$ 0.13\% \\
\cmidrule{2-8}
 & \multirow{7}{*}{2} & 128 & 0.267 & 0.624 & 0.766 & 1.829 & 59.56 $\pm$ 0.19\% \\
 &  & 256 & 0.255  & 0.699 & 0.802 & 1.907 & 60.97 $\pm$ 0.20\% \\
 &  & 512 & 0.247 & 0.746 & 0.829 & 1.729 & 58.19 $\pm$ 0.18\% \\
 &  & 1024 & 0.238 & 0.702 & 0.860 & 1.850 & 50.68 $\pm$ 0.13\% \\
 &  & 2048 & 0.261 & 0.829 & 1.101 & 2.791 & 48.57 $\pm$ 0.22\% \\
 &  & 4096 & 0.266 & 0.944 & 1.451 & 4.822 & 51.79 $\pm$ 0.15\% \\
 &  & 8192 & 0.288 & 1.124 & 2.312 & 9.942 & 60.53 $\pm$ 0.16\% \\
\midrule
Full-context &  & 26031 & - & - & 9.870 & 17.117 & \textbf{72.90 $\pm$ 0.19\%} \\
\midrule
A-Mem &  & 2520 & 0.668 & 1.485 & 1.410 & 4.374 & 48.38 $\pm$ 0.15\% \\
LangMem &  & \textbf{127} & 17.99 & 59.82 & 18.53 & 60.40 & 58.10 $\pm$ 0.21\% \\
Zep &  & 3911 & 0.513 & 0.778 & 1.292 & 2.926 & 65.99 $\pm$ 0.16\% \\
OpenAI &  & 4437 & - & - & 0.466 & 0.889 & 52.90 $\pm$ 0.14\% \\ 
\midrule
\mem &  & 1764 & \textbf{0.148} & \textbf{0.200} & \textbf{0.708} & \textbf{1.440} & 66.88 $\pm$ 0.15\% \\
\memp &  & 3616 & 0.476 & 0.657 & 1.091 & 2.590 & 68.44 $\pm$ 0.17\% \\
\bottomrule
\end{tabular*}
\label{tab:latency_comparison}
\end{table}

In temporal reasoning, \memp~substantially outperforms other methods, validating that structured relational graphs excel in capturing chronological relationships and event sequences. The presence of explicit relational context significantly enhances \memp's temporal coherence, outperforming \mem's dense memory storage and highlighting the importance of precise relational representations when tracking temporally sensitive information. Open-domain performance further reinforces the value of relational modeling. \memp, benefiting from the relational clarity of graph-based memory, closely competes with the top-performing baseline (Zep). This competitive result underscores \memp's robustness in integrating external knowledge through relational clarity, suggesting an optimal synergy between structured memory and open-domain information synthesis.

Overall, our analysis indicates complementary strengths of \mem~and \memp~across various task demands: dense, natural-language-based memory offers significant efficiency for simpler queries, while explicit relational modeling becomes essential for tasks demanding nuanced temporal and contextual integration. 
These findings reinforce the importance of adaptable memory structures tailored to specific reasoning contexts in AI agent deployments.

\begin{figure}[!t]
    \centering
    \begin{subfigure}{\textwidth}
        \centering
        \includegraphics[width=0.97\textwidth]{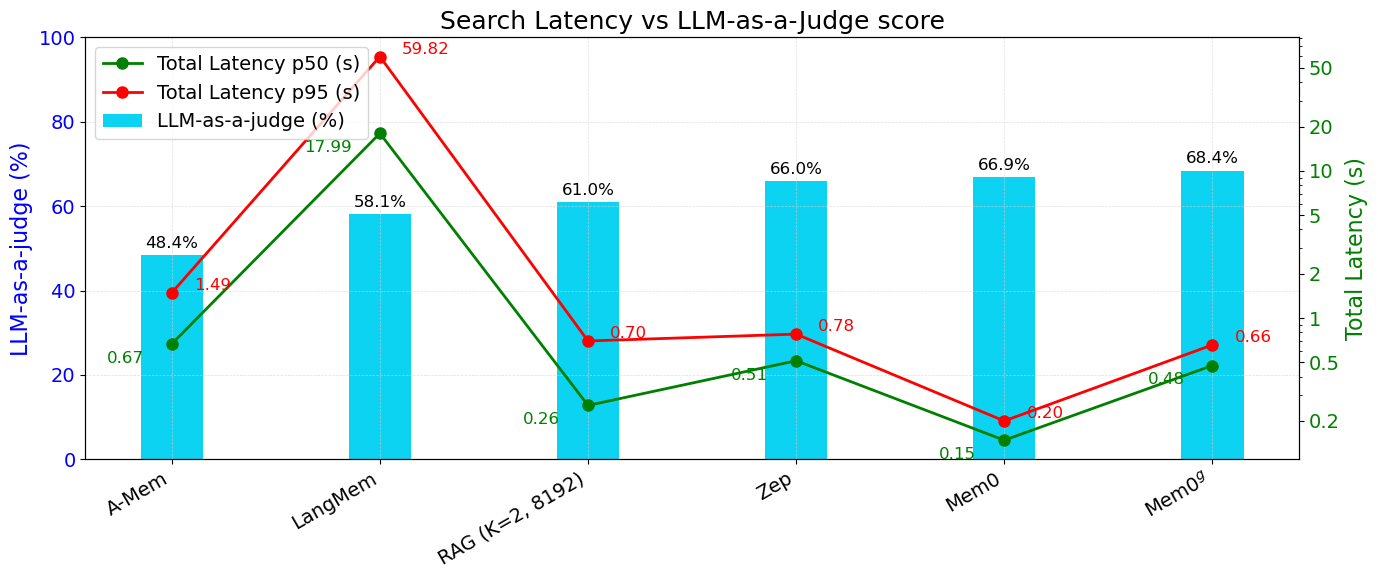}
        \caption{Comparison of \emph{search} latency at p50 (median) and p95 (95th percentile) across different memory methods (\mem, \memp, best RAG variant, Zep, LangMem, and A-Mem). The bar heights represent \Judge~scores (left axis), while the line plots show search latency in seconds (right axis scaled in log).}
        \label{fig:latency_search}
    \end{subfigure}
    \begin{subfigure}{\textwidth}
        \centering
        \includegraphics[width=0.97\textwidth]{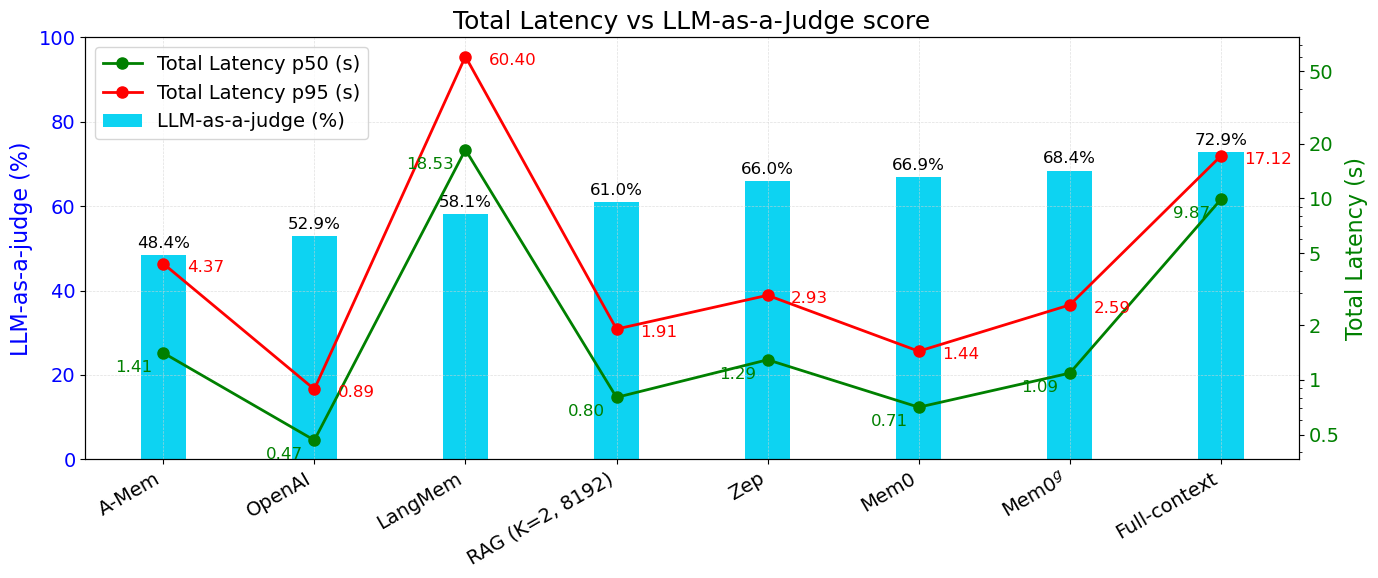}
        \caption{Comparison of \emph{total response} latency at p50 and p95  across different memory methods (\mem, \memp, best RAG variant, Zep, LangMem, OpenAI, full-context, and A-Mem). The bar heights represent \Judge~scores (left axis), and the line plots capture end-to-end latency in seconds (right axis scaled in log).}
        \label{fig:latency_total}
    \end{subfigure}
    \caption{%
    \textbf{Latency Analysis of Different Memory Approaches.} 
    These subfigures illustrate the \Judge~scores and latency comparison of various selected methods from Table \ref{tab:latency_comparison}.
    Subfigure~(a) highlights the \emph{search/retrieval} latency prior to answer generation, while Subfigure~(b) shows the \emph{total} latency (including LLM inference).
    Both plots overlay each method’s \Judge~score for a holistic view of their accuracy and efficiency.
    }
    \label{fig:latency_combined}
\end{figure}

\subsection{Performance Comparison of \mem~and \memp~Against RAG Approaches and Full-Context Model}
Comparisons in Table \ref{tab:latency_comparison}, focusing on the `Overall \Judge' column, reveal that both \mem~and \memp~consistently outperform all RAG configurations, which vary chunk sizes (128–8192 tokens) and retrieve either one ($k$=1) or two ($k$=2) chunks. Even the strongest RAG approach peaks at around 61\% in the \Judge~metric, whereas \mem~reaches 67\%—about a 10\% relative improvement—and \memp~reaches over 68\%, achieving around a 12\% relative gain. These advances underscore the advantage of capturing only the most salient facts in memory, rather than retrieving large chunk of original text. By converting the conversation history into concise, structured representations, \mem~and \memp~mitigate noise and surface more precise cues to the LLM, leading to better answers as evaluated by an external LLM (\Judge).

Despite these improvements, a full-context method that ingests a chunk of roughly 26,000 tokens still achieves the highest \Judge~score (approximately 73\%). However, as shown in Figure \ref{fig:latency_total}, it also incurs a very high total p95 latency—around 17 seconds—since the model must read the entire conversation on every query. By contrast, \mem~and \memp~significantly reduce token usage and thus achieve lower p95 latencies of around 1.44 seconds (a 92\% reduction) and 2.6 seconds (a 85\% reduction), respectively over full-context approach. Although the full-context approach can provide a slight accuracy edge, the memory-based systems offer a more practical trade-off, maintaining near-competitive quality while imposing only a fraction of the token and latency cost. 
As conversation length increases, full-context approaches suffer from exponential growth in computational overhead (evident in Table \ref{tab:latency_comparison} where total p95 latency increases significantly with larger $k$ values or chunk sizes). This increase in input chunks leads to longer response times and higher token consumption costs. In contrast, memory-focused approaches like \mem~and \memp~maintain consistent performance regardless of conversation length, making them substantially more viable for production-scale deployments where efficiency and responsiveness are critical.

\subsection{Latency Analysis}
Table \ref{tab:latency_comparison} provides a comprehensive performance comparison of various retrieval and memory methodologies, presenting median (p50) and tail (p95) latencies for both the search phase and total response generation across the \texttt{LOCOMO} dataset. Our analysis reveals distinct performance patterns governed by architectural choices. Memory-centric architectures demonstrate different performance characteristics. A-Mem, despite its larger memory store, incurs substantial search overhead (p50: 0.668s), resulting in total median latencies of 1.410s. LangMem exhibits even higher search latencies (p50: 17.99s, p95: 59.82s), rendering it impractical for interactive applications. Zep achieves moderate performance (p50 total: 1.292s).
The full-context baseline, which processes the entire conversation history without retrieval, fundamentally differs from retrieval-based approaches. By passing the entire conversation context (26000 tokens) directly to the LLM, it eliminates search overhead but incurs extreme total latencies (p50: 9.870s, p95: 17.117s). Similarly, the OpenAI implementation does not perform memory search, as it processes manually extracted memories from their playground. While this approach achieves impressive response generation times (p50: 0.466s, p95: 0.889s), it requires pre-extraction of relevant context, which is not reflected in the reported metrics.

Our proposed \mem~approach achieves the lowest search latency among all methods (p50: 0.148s, p95: 0.200s) as illustrated in Figure \ref{fig:latency_search}. This efficiency stems from our selective memory retrieval mechanism and infra improvements that dynamically identifies and retrieves only the most salient information rather than fixed-size chunks. Consequently, \mem~maintains the lowest total median latency (0.708s) with remarkably contained p95 values (1.440s), making it particularly suitable for latency-sensitive applications such as interactive AI agents. The graph-enhanced \memp~variant introduces additional relational modeling capabilities at a moderate latency cost, with search times (0.476s) still outperforming all existing memory solutions and baselines. 
Despite this increase, \memp~maintains competitive total latencies (p50: 1.091s, p95: 2.590s) while achieving the highest \Judge~score (68.44\%) across all methods—trailing only the computationally prohibitive full-context approach.
This performance profile demonstrates our methods' ability to balance response quality and computational efficiency, offering a compelling solution for production AI agents where both factors are critical constraints.

\subsection{Memory System Overhead: Token Analysis and Construction Time}
We measure the average token budget required to materialise each system's long‑term memory store.
\mem~encodes complete dialogue turns in a natural language representation and therefore occupies only \textbf{7k} tokens per conversation on an average. Where as \memp~roughly doubles the footprint to \textbf{14k} tokens, due to the introduction of graph memories which includes nodes and corresponding relationships. In stark contrast, Zep's memory graph consumes in excess of \textbf{600k} tokens. The inflation arises from Zep's design choice to cache a full abstractive summary at every node while also storing facts on the connecting edges, leading to extensive redundancy across the graph. For perspective, supplying the \emph{entire} raw conversation context to the language model—without any memory abstraction—amounts to roughly \textbf{26k} tokens on average, 20 times less relative to Zep's graph.
Beyond token inefficiency, our experiments revealed significant operational bottlenecks with Zep. After adding memories to Zep's system, we observed that immediate memory retrieval attempts often failed to answer our queries correctly. Interestingly, re-running identical searches after a delay of several hours yielded considerably better results. This latency suggests that Zep's graph construction involves multiple asynchronous LLM calls and extensive background processing, making the memory system impractical for real-time applications. In contrast, \mem~graph construction completes in under a minute even in worst-case scenarios, allowing users to immediately leverage newly added memories for query responses.

These findings highlight that Zep not only replicates identical knowledge fragments across multiple nodes, but also introduces significant operational delays. Our architectures—\mem~and \memp—preserve the same information at a fraction of the token cost and with substantially faster memory availability, offering a more memory‑efficient and operationally responsive representation.

%% file: sections/conclusion.tex
\section{Conclusion and Future Work}
\label{sec:conclusion}

We have introduced \mem~and \memp, two complementary memory architectures that overcome the intrinsic limitations of fixed context windows in LLMs. By dynamically extracting, consolidating, and retrieving compact memory representations, \mem~achieves state-of-the-art performance across single‑hop and multi‑hop reasoning, while \memp’s graph‑based extensions unlock significant gains in temporal and open‑domain tasks. On the \texttt{LOCOMO} benchmark, our methods deliver 5\%, 11\%, and 7\% relative improvements in single-hop, temporal, and multi-hop reasoning question types over best performing methods in respective question type and reduce p95 latency by over 91\% compared to full‑context baselines—demonstrating a powerful balance between precision and responsiveness. \mem’s dense memory pipeline excels at rapid retrieval for straightforward queries, minimizing token usage and computational overhead. In contrast, \memp’s structured graph representations provide nuanced relational clarity, enabling complex event sequencing and rich context integration without sacrificing practical efficiency. Together, they form a versatile memory toolkit that adapts to diverse conversational demands while remaining deployable at scale.

Future research directions include optimizing graph operations to reduce the latency overhead in \memp, exploring hierarchical memory architectures that blend efficiency with relational representation, and developing more sophisticated memory consolidation mechanisms inspired by human cognitive processes. Additionally, extending our memory frameworks to domains beyond conversational scenarios, such as procedural reasoning and multimodal interactions, would further validate their broader applicability. By addressing the fundamental limitations of fixed context windows, our work represents a significant advancement toward conversational AI systems capable of maintaining coherent, contextually rich interactions over extended periods, much like their human counterparts.

\section{Acknowledgments}
We would like to express our sincere gratitude to Harsh Agarwal, Shyamal Anadkat, Prithvijit Chattopadhyay, Siddesh Choudhary, Rishabh Jain, and Vaibhav Pandey for their invaluable insights and thorough reviews of early drafts. Their constructive comments and detailed suggestions helped refine the manuscript, enhancing both its clarity and overall quality. We deeply appreciate their generosity in dedicating time and expertise to this work.

%% file: sections/appendix.tex
\newpage
\appendix

\section*{\center\LARGE Appendix}

\vspace{4mm}

\section{Prompts}
\label{appendix:llm_judge}
In developing our LLM-as-a-Judge prompt, we adapt elements from the prompt released by \cite{packer2023memgpt}.

\NewTColorBox{EqBox}{ s O {!htbp} m }{%
  floatplacement={#2},
  IfBooleanTF={#1}{float*,width=\textwidth}{float},
  title={\textsc{#3}},
}

\begin{EqBox}[!htbp]{Prompt Template for LLM as a Judge}
\vspace{1mm}
{\tt \footnotesize  
Your task is to label an answer to a question as "CORRECT" or "WRONG". 
You will be given the following data:
    (1) a question (posed by one user to another user), 
    (2) a `gold' (ground truth) answer, 
    (3) a generated answer
which you will score as CORRECT/WRONG.

\vspace{0.5cm}
The point of the question is to ask about something one user should know about the other user based on their prior conversations.
The gold answer will usually be a concise and short answer that includes the referenced topic, for example:

Question: Do you remember what I got the last time I went to Hawaii?

Gold answer: A shell necklace

The generated answer might be much longer, but you should be generous with your grading - as long as it touches on the same topic as the gold answer, it should be counted as CORRECT. 

\vspace{0.5cm}
For time related questions, the gold answer will be a specific date, month, year, etc. The generated answer might be much longer or use relative time references (like `last Tuesday' or `next month'), but you should be generous with your grading - as long as it refers to the same date or time period as the gold answer, it should be counted as CORRECT. Even if the format differs (e.g., `May 7th' vs `7 May'), consider it CORRECT if it's the same date.

\vspace{0.5cm}
Now it's time for the real question:

Question: \{question\}

Gold answer: \{gold\_answer\}

Generated answer: \{generated\_answer\}

\vspace{0.5cm}
First, provide a short (one sentence) explanation of your reasoning, then finish with CORRECT or WRONG. 
Do NOT include both CORRECT and WRONG in your response, or it will break the evaluation script.

\vspace{0.5cm}
Just return the label CORRECT or WRONG in a json format with the key as "label".
}
\end{EqBox}

\begin{EqBox}[!htbp]{Prompt Template for Results Generation (\mem)}
\vspace{1mm}
{\tt \footnotesize  
You are an intelligent memory assistant tasked with retrieving accurate information from conversation memories. \vspace{0.5cm}

\# CONTEXT: \vspace{0.5cm}

You have access to memories from two speakers in a conversation. These memories contain timestamped information that may be relevant to answering the question.
\vspace{0.5cm}

\# INSTRUCTIONS: \vspace{0.5cm}

1. Carefully analyze all provided memories from both speakers

2. Pay special attention to the timestamps to determine the answer

3. If the question asks about a specific event or fact, look for direct evidence in the memories

4. If the memories contain contradictory information, prioritize the most recent memory

5. If there is a question about time references (like "last year", "two months ago", etc.), 
   calculate the actual date based on the memory timestamp. For example, if a memory from 
   4 May 2022 mentions "went to India last year," then the trip occurred in 2021.
   
6. Always convert relative time references to specific dates, months, or years. For example, 
   convert "last year" to "2022" or "two months ago" to "March 2023" based on the memory 
   timestamp. Ignore the reference while answering the question.
   
7. Focus only on the content of the memories from both speakers. Do not confuse character 
   names mentioned in memories with the actual users who created those memories.
   
8. The answer should be less than 5-6 words.

\vspace{0.5cm}
\# APPROACH (Think step by step): \vspace{0.5cm}

1. First, examine all memories that contain information related to the question

2. Examine the timestamps and content of these memories carefully

3. Look for explicit mentions of dates, times, locations, or events that answer the question

4. If the answer requires calculation (e.g., converting relative time references), show your work

5. Formulate a precise, concise answer based solely on the evidence in the memories

6. Double-check that your answer directly addresses the question asked

7. Ensure your final answer is specific and avoids vague time references

\vspace{0.5cm}
Memories for user \{speaker\_1\_user\_id\}:

\{speaker\_1\_memories\}
\vspace{0.5cm}

Memories for user \{speaker\_2\_user\_id\}:

\{speaker\_2\_memories\}

\vspace{0.5cm}
Question: \{question\}

\vspace{0.5cm}
Answer:
}
\end{EqBox}

\begin{EqBox}[!htbp]{Prompt Template for Results Generation (\memp)}
\vspace{1mm}
{\tt \footnotesize  
(\textit{same as previous})
\vspace{0.5cm}

\# APPROACH (Think step by step):\vspace{0.5cm}

1. First, examine all memories that contain information related to the question

2. Examine the timestamps and content of these memories carefully

3. Look for explicit mentions of dates, times, locations, or events that answer the 
   question
   
4. If the answer requires calculation (e.g., converting relative time references), 
   show your work
   
5. Analyze the knowledge graph relations to understand the user's knowledge context

6. Formulate a precise, concise answer based solely on the evidence in the memories

7. Double-check that your answer directly addresses the question asked

8. Ensure your final answer is specific and avoids vague time references

\vspace{0.5cm}
Memories for user \{speaker\_1\_user\_id\}:

\{speaker\_1\_memories\}
\vspace{0.5cm}

Relations for user \{speaker\_1\_user\_id\}:

\{speaker\_1\_graph\_memories\}
\vspace{0.5cm}

Memories for user \{speaker\_2\_user\_id\}:

\{speaker\_2\_memories\}
\vspace{0.5cm}

Relations for user \{speaker\_2\_user\_id\}:

\{speaker\_2\_graph\_memories\}
\vspace{0.5cm}

Question: \{question\}
\vspace{0.5cm}

Answer:
}
\end{EqBox}

\begin{EqBox}[!htbp]{Prompt Template for OpenAI ChatGPT}
\vspace{1mm}
{\tt \footnotesize  
Can you please extract relevant information from this conversation and create memory entries for each user mentioned? Please store these memories in your knowledge base in addition to the timestamp provided for future reference and personalized interactions.
\vspace{0.5cm}

(1:56 pm on 8 May, 2023) Caroline: Hey Mel! Good to see you! How have you been?

(1:56 pm on 8 May, 2023) Melanie: Hey Caroline! Good to see you! I'm swamped with the kids \& work. What's up with you? Anything new?

(1:56 pm on 8 May, 2023) Caroline: I went to a LGBTQ support group yesterday and it was so powerful.

...
}
\end{EqBox}

\clearpage

\section{Algorithm}
\label{appendix:algorithm}

\begin{algorithm}[!h]
\caption{Memory Management System: Update Operations}
\label{alg:memory_update}
\begin{algorithmic}[1]
\smallskip
\State \textbf{Input:} Set of retrieved memories $F$, Existing memory store $M = \{m_1, m_2, \ldots, m_n\}$
\State \textbf{Output:} Updated memory store $M'$
\smallskip
\hrule
\smallskip
\Procedure{UpdateMemory}{$F, M$}
    \For{each fact $f \in F$}
        \State $operation \gets \Call{ClassifyOperation}{f, M}$
        \smallskip
        \Comment{Execute appropriate operation based on classification}
        \If{$operation = \text{ADD}$}
            \State $id \gets \text{GenerateUniqueID}()$
            \State $M \gets M \cup \{(id, f, \text{"ADD"})\}$
            \Comment{Add new fact with unique identifier}
        \ElsIf{$operation = \text{UPDATE}$}
            \State $m_i \gets \text{FindRelatedMemory}(f, M)$
            \If{$\text{InformationContent}(f) > \text{InformationContent}(m_i)$}
                \State $M \gets (M \setminus \{m_i\}) \cup \{(id_i, f, \text{"UPDATE"})\}$
                \Comment{Replace with richer information}
            \EndIf
        \ElsIf{$operation = \text{DELETE}$}
            \State $m_i \gets \text{FindContradictedMemory}(f, M)$
            \State $M \gets M \setminus \{m_i\}$
            \Comment{Remove contradicted information}
        \ElsIf{$operation = \text{NOOP}$}
            \State \text{No operation performed}
            \Comment{Fact already exists or is irrelevant}
        \EndIf
        \smallskip
    \EndFor
    \State \Return $M$
\EndProcedure
\smallskip
\hrule
\smallskip
\Function{ClassifyOperation}{$f, M$}
    \If{$\lnot \text{SemanticallySimilar}(f, M)$}
        \State \Return \text{ADD}
        \Comment{New information not present in memory}
    \ElsIf{$\text{Contradicts}(f, M)$}
        \State \Return \text{DELETE}
        \Comment{Information conflicts with existing memory}
    \ElsIf{$\text{Augments}(f, M)$}
        \State \Return \text{UPDATE}
        \Comment{Enhances existing information in memory}
    \Else
        \State \Return \text{NOOP}
        \Comment{No change required}
    \EndIf
\EndFunction
\end{algorithmic}
\end{algorithm}

\section{Selected Baselines}
\label{appendix:baselines}

\paragraph{LoCoMo} The LoCoMo framework implements a sophisticated memory pipeline that enables LLM agents to maintain coherent, long-term conversations. At its core, the system divides memory into short-term and long-term components. After each conversation session, agents generate summaries (stored as short-term memory) that distill key information from that interaction. Simultaneously, individual conversation turns are transformed into `\textit{observations}' - factual statements about each speaker's persona and life events that are stored in long-term memory with references to the specific dialog turns that produced them. When generating new responses, agents leverage both the most recent session summary and selectively retrieve relevant observations from their long-term memory. This dual-memory approach is further enhanced by incorporating a temporal event graph that tracks causally connected life events occurring between conversation sessions. By conditioning responses on retrieved memories, current conversation context, persona information, and intervening life events, the system enables agents to maintain consistent personalities and recall important details across conversations spanning hundreds of turns and dozens of sessions.

\paragraph{ReadAgent} ReadAgent addresses the fundamental limitations of LLMs by emulating how humans process lengthy texts through a sophisticated three-stage pipeline. First, in Episode Pagination, the system intelligently segments text at natural cognitive boundaries rather than arbitrary cutoffs. Next, during Memory Gisting, it distills each segment into concise summaries that preserve essential meaning while drastically reducing token count—similar to how human memory retains the substance of information without verbatim recall. Finally, when tasked with answering questions, the Interactive Lookup mechanism examines these gists and strategically retrieves only the most relevant original text segments for detailed processing. This human-inspired approach enables LLMs to effectively manage documents up to 20 times longer than their normal context windows. By balancing global understanding through gists with selective attention to details, ReadAgent achieves both computational efficiency and improved comprehension, demonstrating that mimicking human cognitive processes can significantly enhance AI text processing capabilities.

\paragraph{MemoryBank} The MemoryBank system enhances LLMs with long-term memory through a sophisticated three-part pipeline. At its core, the Memory Storage component warehouses detailed conversation logs, hierarchical event summaries, and evolving user personality profiles. When a new interaction occurs, the Memory Retrieval mechanism employs a dual-tower dense retrieval model to extract contextually relevant past information. The Memory Updating component, provides a human-like forgetting mechanism where memories strengthen when recalled and naturally decay over time if unused. This comprehensive approach enables AI companions to recall pertinent information, maintain contextual awareness across extended interactions, and develop increasingly accurate user portraits, resulting in more personalized and natural long-term conversations.

\paragraph{MemGPT} The MemGPT system introduces an operating system-inspired approach to overcome the context window limitations inherent in LLMs. At its core, MemGPT employs a sophisticated memory management pipeline consisting of three key components: a hierarchical memory system, self-directed memory operations, and an event-based control flow mechanism. The system divides available memory into `\textit{main context}' (analogous to RAM in traditional operating systems) and `\textit{external context}' (analogous to disk storage). The main context—which is bound by the LLM's context window—contains system instructions, recent conversation history, and working memory that can be modified by the model. The external context stores unlimited information outside the model's immediate context window, including complete conversation histories and archival data. When the LLM needs information not present in main context, it can initiate function calls to search, retrieve, or modify content across these memory tiers, effectively `\textit{paging}' relevant information in and out of its limited context window. This OS-inspired architecture enables MemGPT to maintain conversational coherence over extended interactions, manage documents that exceed standard context limits, and perform multi-hop information retrieval tasks—all while operating with fixed-context models. The system's ability to intelligently manage its own memory resources provides the illusion of infinite context, significantly extending what's possible with current LLM technology.

\paragraph{A-Mem} The A-Mem model introduces an agentic memory system designed for LLM agents. This system dynamically structures and evolves memories through interconnected notes. Each note captures interactions enriched with structured attributes like keywords, contextual descriptions, and tags generated by the LLM. Upon creating a new memory, A-MEM uses semantic embeddings to retrieve relevant existing notes, then employs an LLM-driven approach to establish meaningful links based on similarities and shared attributes. Crucially, the memory evolution mechanism updates existing notes dynamically, refining their contextual information and attributes whenever new relevant memories are integrated. Thus, memory structure continually evolves, allowing richer and contextually deeper connections among memories. Retrieval from memory is conducted through semantic similarity, providing relevant historical context during agent interactions